\documentclass[letterpaper, 10 pt, conference]{ieeeconf}  % Comment this line out if you need a4paper

\IEEEoverridecommandlockouts                              % This command is only needed if 
\usepackage[final]{graphicx}
\usepackage[T1]{fontenc}
\usepackage{float}
\usepackage{flushend}
\usepackage[colorlinks = true,linkcolor = blue,urlcolor  = blue,]{hyperref}
\usepackage{xcolor}
\newcommand{\cmt}[1]{}

\long\def\ignorethis#1{}

\newcommand{\etal}{{\em{et~al.}\ }}
\newcommand{\eg}{e.g.\ }

\newcommand{\figref}[1]{Fig.~\ref{fig:#1}}
\newcommand{\pctab}{\hspace{0.2in}}

\title{\LARGE \bf
Learning to Navigate Sidewalks in Outdoor Environments
}

\author{
 Maks Sorokin$^{1}$, Jie Tan$^{2}$, C. Karen Liu$^{3}$, Sehoon Ha$^{12}$
 
\thanks{$^{1}$ Georgia Institute of Technology, Atlanta, GA, 30308, USA 
}
\thanks{
\hspace{2mm}\tt\small \{maks,sehoonha\}@gatech.edu
}

\thanks{$^{2}$ Robotics at Google, Mountain View, CA, 94043, USA
}
\thanks{
\hspace{2mm}
\tt\small jietan@google.com
}

\thanks{$^{3}$ Stanford University, Stanford,CA, 94305, USA
}
\thanks{
\hspace{2mm}
\tt\small karenliu@cs.stanford.edu
}

}

\begin{document}

\maketitle

\thispagestyle{empty}
\pagestyle{empty}

%%%%%%%%%%%%%%%%%%%%%%%%%%%%%%%%%%%%%%%%%%%%%%%%%%%%%%%%%%%%%%%%%%%%%%%%%%%%%%
\begin{abstract}
Outdoor navigation on sidewalks in urban environments is the key technology behind important human assistive applications, such as last-mile delivery or neighborhood patrol. This paper aims to develop a quadruped robot that follows a route plan generated by public map services, while remaining on sidewalks and avoiding collisions with obstacles and pedestrians. We devise a two-staged learning framework, which first trains a teacher agent in an abstract world with privileged ground-truth information, and then applies Behavior Cloning to teach the skills to a student agent who only has access to realistic sensors. The main research effort of this paper focuses on overcoming challenges when deploying the student policy on a quadruped robot in the real world. We propose methodologies for designing sensing modalities, network architectures, and training procedures to enable zero-shot policy transfer to unstructured and dynamic real outdoor environments. We evaluate our learning framework on a quadrupedal robot navigating sidewalks in the city of Atlanta, USA. Using the learned navigation policy and its onboard sensors, the robot is able to walk 3.2 kilometers with a limited number of human interventions.

\textbf{Project webpage: \href{https://initmaks.com/navigation}{https://initmaks.com/navigation}}
\end{abstract}
%%%%%%%%%%%%%%%%%%%%%%%%%%%%%%%%%%%%%%%%%%%%%%%%%%%%%%%%%%%%%%%%%%%%%%%%%%%%%%

\section{Introduction} \label{sec:intro}

Outdoor navigation by foot in an urban environment is an essential life skill we acquired at a young age. Likewise, autonomous robotic workers that assist humans in an urban environment, such as last-mile delivery or neighborhood patrol, also need to learn to navigate sidewalks and avoid collisions. Similar to autonomous driving and indoor navigation, sidewalk navigation requires the robots to follow a route plan under noisy localization signals and egocentric sensors. However, outdoor sidewalk navigation faces more unstructured environments with a wide variety of pedestrians and obstacles and without any guiding lanes.

This paper aims to develop a learning framework that allows a quadrupedal robot to follow a route plan generated by public map services, such as Google or Apple Maps, while remaining on sidewalks and avoiding collisions. 
We take a two-staged learning approach similar to "learning by cheating" \cite{chen2020learning}, where we first train a teacher agent in an abstract world with privileged information of bird-eye-view observations. Then clone the learned behavior to a student policy with realistic sensor configurations using Dataset Aggregation Method (DAGGER)~\cite{ross2011reduction}.

\begin{figure}
    \centering
    \includegraphics[width=0.99\linewidth]{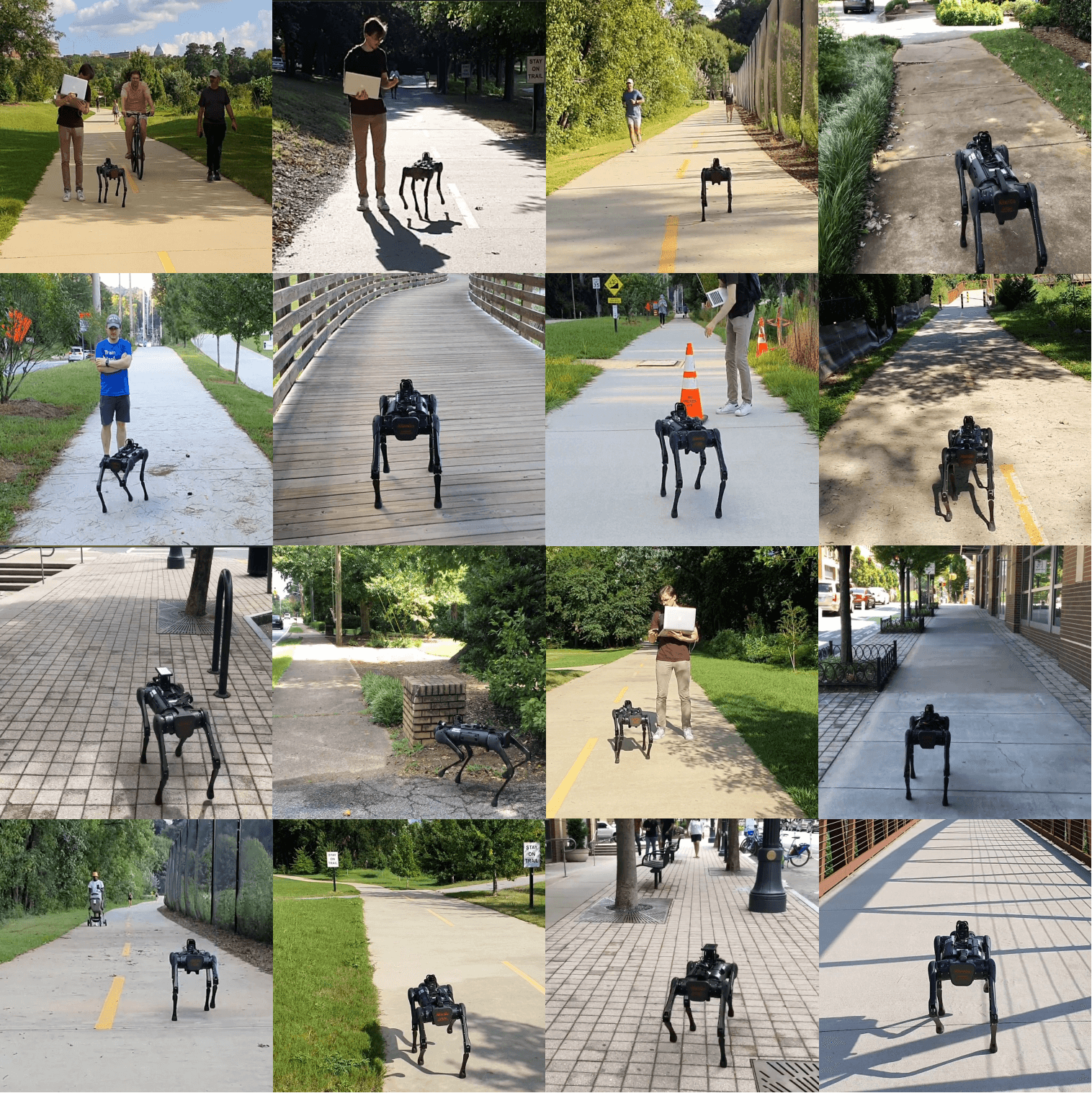}
    \caption{AlienGo robot navigating various real-world sidewalks in outdoor.}
    \label{fig:teaser}
\end{figure}

However, directly applying "learning by cheating" \cite{chen2020learning} to outdoor navigation problem results in a poor policy that fails immediately in the real world. Therefore, the main research contribution of this work is to methodologically ablate the sim-to-real gap and find pragmatic solutions to mitigate the primary sources of the gap, enabling \cite{chen2020learning} to be applied to the real robots. 

Guided by careful analyses of the sim-to-real gap, we propose new methodologies for designing sensing modalities, network architectures, and training procedures. These inventions allow us to achieve the challenging goal of quadrupeds navigating outdoor sidewalks for a long distance with only a limited number of human interventions and without any curation or arrangement of the environment.

We evaluate the proposed framework on a real quadrupedal robot, AlienGo from Unitree, to walk the sidewalks in the city of Atlanta, USA. In our experiments, the robot navigated the total $3.2$~kilometers of sidewalks using egocentric camera, LiDAR, and GPS with a few human interventions. Our natural testing environments include diverse static and dynamic objects typically seen in an urban space, such as sidewalks, trees, bridges, pedestrians, dogs, and bicycle riders. We also validate our framework by comparing different learning algorithms and sensor configurations in both simulated and real-world environments. Our technical contributions include:
\begin{itemize}
\item We develop a two-staged learning framework that leverages two different simulators, inspired by the ``learning by cheating'' approach.
\item We conduct a careful analysis of the \emph{sim-to-real gap} and propose design choices and mitigation techniques.
\item We demonstrate real-world outdoor sidewalk navigation on a quadrupedal robot, which spans $3.2$ kilometers.
\end{itemize}

\section{Related Work} \label{sec:related}

% Our work is built on previous work on visual navigation, teacher-student learning, and sim-to-real techniques.

\subsection{Visual Navigation}
Researchers have studied vision-based robot navigation for decades using both hand-engineered and learning-based approaches. In 2007, Morales \etal\cite{morales2009autonomous} presented an autonomous robot system that can traverse previously mapped sidewalk trajectories in the Tsukuba Challenge. Kümmerle \etal\cite{kummerle2015autonomous} showed autonomous navigation of a $3$~km route in the city center of Freiburg using the pre-built map data and a large array of laser sensors. While demonstrating impressive results, these systems require extensive manual engineering of various components, including sensor configuration, calibration, signal processing, and sensor fusion. As reported in their papers, it is also not straightforward to achieve complex behaviors, such as pedestrian avoidance.

On the other hand, deep learning enables us to learn effective navigation policies from a large amount of experience. The recent development of high-performance navigation simulators, such as Habitat~\cite{habitat19iccv} and iGibson~\cite{shen2021igibson}, has enabled researchers to develop large scale visual navigation algorithms~\cite{gupta2017cognitive,fang2019scene,wijmans2019dd,chaplot2020learning,francis2020long,petrenko2020sample,hoeller2021learning,karnan2021voila} for indoor environments. 
Wijmans \etal~\cite{wijmans2019dd} proposed an end-to-end vision-based reinforcement learning algorithm to train near-perfect agents that can navigate unseen indoor environments without access to the map by leveraging billions of simulation samples. Chaplot \etal~\cite{chaplot2020learning} presented a hierarchical and modular approach for indoor floor construction \& floor exploration that combines both learning and traditional algorithms. Francis \etal~\cite{francis2020long} showed an effective indoor navigation skill solely based on a single laser sensor.

While a lot of progress in learned indoor navigation has been made, only a handful of learning algorithms have been developed for outdoor navigation~\cite{muller2018driving,kahn2021badgr,kahn2021land}. Müller \etal~\cite{muller2018driving} trained a waypoint navigation policy in simulation and deployed it on a real toy-sized truck to drive on an empty vehicle road. Kahn \etal~\cite{kahn2021land} proposed an approach for learning a sidewalk following policy purely in the real world by leveraging human-operator disengagements and overtakes. Instead, we demonstrate that it is possible to learn a policy to follow a designated set of waypoints on sidewalks without relying on any real-world data.

\subsection{Teacher-Student Framework}

While end-to-end learning is a viable method, some challenging tasks require more structured learning algorithms, particularly when they involve large observation or state spaces that are hard to explore. Recently, Chen \etal~\cite{chen2020learning} proposed the  ``\emph{learning by cheating}'' framework that can effectively learn complex policies by taking a two-staged approach. It first learns an expert policy that ``cheats'' by accessing privileged ground-truth information and transferring the learned behaviors with realistic sensor modalities. Lee \etal~\cite{lee2020learning} demonstrated that this framework can learn a robust locomotion policy on challenging terrains.

We use the ``\emph{learning by cheating}'' framework as a fundamental building block and demonstrate how we can further accelerate it by leveraging fast-speed simulation environments~\cite{petrenko2020sample,petrenko2021megaverse}.  While the original paper did not address the sim-to-real transfer, with our proposed techniques, we show that learning by cheating framework can be extended to real quadrupeds navigating in well-populated urban areas.

\subsection{Sim-to-Real Transfer}
Overcoming the sim-to-real gap has been an active research topic in recent years~\cite{sadeghi2016cad2rl,tobin2017domain,muller2018driving,james2019sim}. For instance, previous work~\cite{sadeghi2016cad2rl,tobin2017domain} investigated the Domain Randomization approach by applying random textures to the world surfaces to learn texture agnostic visual features. James \etal~\cite{james2019sim} proposed to train generative adversarial networks to reduce visual discrepancies between simulation and real-world images. Laskin \etal~\cite{laskin2020reinforcement} proposed an image-based data augmentation technique to avoid overfitting, which applies random transforms to image observations.

\begin{figure*}
    \centering
    \vspace{-0.7cm}
    \includegraphics[width=0.95\linewidth]{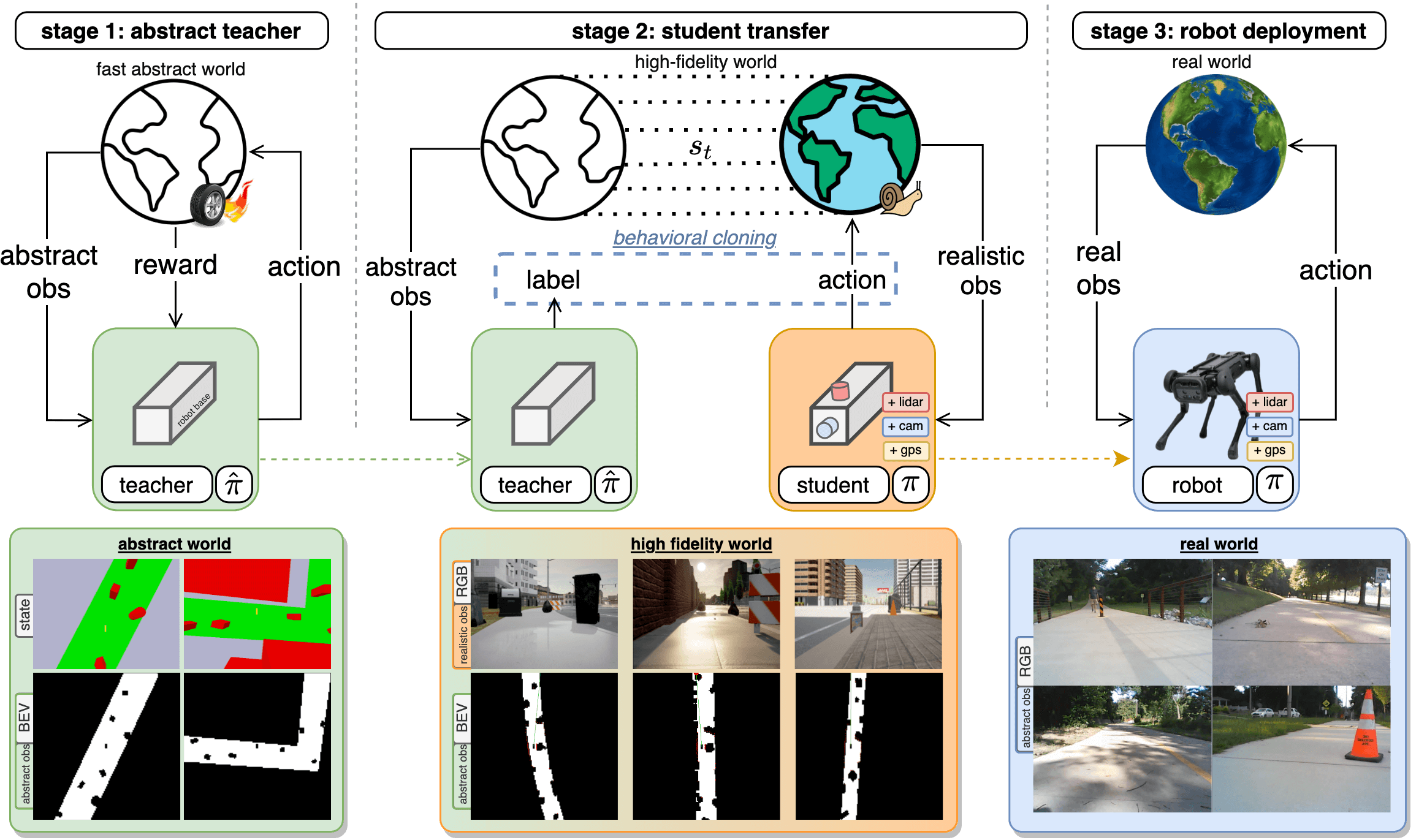}
    \caption{Overview of the entire pipeline from abstract to real world deployment. A teacher is trained in an abstract world with privileged sensing information but simpler geometry and rendering thus enabling faster learning. A student is cloned in a high-fidelity simulation with realistic sensors, and finally deployed to the real world. 
    }
    \label{fig:overview}
\end{figure*}

Inspired by the work of Müller \etal~\cite{muller2018driving}, we also employ pretrained semantic networks for visual inference to help the sim-to-real transfer. However,  we found that learned semantics networks with publicly available autonomous driving data sets show poor prediction on sidewalk navigation due to the perspective shift. Therefore, we retrain semantics with more sidewalk perspective images collected from virtual worlds~\cite{richter2017playing}. We also train a policy using intermediate semantic features, as Shah \etal~\cite{shah2021rrl} suggested. Combining these techniques, we were able to bridge the visual sim-to-real gap for sidewalk navigation effectively.
\section{Navigation policy} \label{sec:navpi}

This section describes a learning algorithm to obtain a visual navigation policy for outdoor sidewalk navigation. Because of the complexity of the given problem, we adopt an approach of ``learning by cheating''~\cite{chen2020learning,lee2020learning}, that decomposes the problem into learning of teacher and student policies (\figref{overview}). First, we perform teacher training in an abstract world with salient information most relevant to the task of navigation. We use a birds-eye-view image as the privileged observation and efficiently learn a navigation policy. Once we obtain satisfactory performance, we train a student agent which only has access to realistic sensors (hence more limited sensing capability) in a high-fidelity simulator. We use DAGGER~\cite{ross2011reduction} train the student policy. This two-staged learning allows us to train a policy much more efficiently than learning it from scratch.

\subsection{Teacher Training in Abstract World} \label{abstract_sim}

In the ``learning by cheating'' framework, teacher learning aims to efficiently obtain the ideal control policy by allowing it to access ``privileged'' information that cannot be obtained with robot's onboard sensors.  In our scenario, privileged information is a birds-eye-view image that captures map layouts and nearby obstacles.

Because a high-fidelity simulator is computationally costly, we accelerate the learning by employing a simple abstract world for teacher training. This abstract world only contains essential information required for sidewalk navigation, such as walkable/non-walkable area, and static/dynamic objects (\figref{overview}). We create this abstract world using PyBullet~\cite{coumans2021}, which provides a fast and simple interface for accessing and rendering the abstract world observation modalities. In our experience, an abstract world can generate samples more than $10$ times faster than a high-fidelity simulator, CARLA~\cite{dosovitskiy2017carla} because it has simplified geometry and avoids heavy rendering pipelines.

\noindent \textbf{Environment.} We generate the abstract world using the actual city data of the OpenStreetMap~\cite{OpenStreetMap} with the help of OSM2World~\cite{OSM2World}. We parse the map layout information of the Helsinki area around $1km^2$ in size. To make the simulated empty street scenes closer to those in the real world, we randomly populate sidewalks with cylinder and cuboid-shaped objects.
% \sehoon{can they be dynamic?} \maks{they can be, even used to be, but aren't in the current setup.}
We also vary the sidewalk width from $2$ to $5$ meters for a richer training experience.

\noindent \textbf{Task.}
The main objective of the teacher agent is to navigate the sidewalk to follow a list of waypoints on the map while avoiding collisions and respecting the sidewalk boundary. For each episode, we randomly select the start and goal sidewalk positions $10$ to $15$ meters apart while guaranteeing reachability. We design the reward function similar to other navigation papers~\cite{savva2019habitat}:
$$r = r_{success} + r_{termination} + r_{approach} + r_{life}.$$
$r_{success}$ is a sparse reward of $10.0$ awarded if the agents gets close to the goal (< $0.5$m).
$r_{termination}$ is another sparse term of $-10.0$ if the agent collides with any obstacle, moves off the sidewalk, or fails to reach the goal within 150 simulation steps.
$r_{approach}$ acts as a continuous incentive for making progress defined by the difference between the previous and the current distances to the goal.
$r_{life}$ a small life penalty of $-0.01$ to avoid the agent from idling.

\noindent \textbf{Observation Space.}
We define a \emph{abstract} observation space (\figref{overview}) with privileged information, which can be obtained from both the abstract and high-fidelity worlds. It consists of the following components:

\begin{enumerate}

\item \textbf{Bird-Eye-View Image} (BEV, privileged) is a 128x128 top-down binary image which covers a $18$m by $18$m region around the robot. Each pixel is one if the region is walkable while 0 for objects, walls, buildings, etc. We stack three previous BEVs to consider the history.

\item \textbf{Bird-Eye-View Lidar} (BLID) is a simulated LiDAR observation that measures distances to the nearest objects by casting $64$ rays.

\item \textbf{Goal direction \& distance} (GDD) is a relative goal location in the local polar coordinate frame.

\end{enumerate}

\noindent \textbf{Action Space.} Each action is defined with two continuous values. The first action controls the speed ($20$~cm to $-10$~cm per step) in the forward/backward direction, while the second action controls the rotation ($-54^\circ$ to $54^\circ$ per step) around the yaw axis.

\noindent \textbf{Learning.}
To train the teacher agent, we use a modification of the Soft Actor-Critic~\cite{haarnoja2018soft,yarats2019improving}. We adjust the Actor and Critic networks to incorporate the BLID and GDD observations by stacking them with features extracted by a visual encoder.

We train the policy until saturation, which reaches the success rate of $83\%$ on the validation environments in the high fidelity world.

\subsection{Student Training in High fidelity world} \label{carla_world}

Once we obtain an effective teacher policy, the next step is to learn a student policy in a high-fidelity simulator with realistic observations that the robot's onboard sensors can provide. We train a student agent by cloning teacher's behavior using Dataset Aggregation Method (DAGGER)~\cite{ross2011reduction}. 

\noindent \textbf{Environment.}
We train a student policy in CARLA~\cite{dosovitskiy2017carla} (\figref{overview}), a high fidelity simulator built on the Unreal Engine 4. The simulator includes multiple town blocks with various weather conditions, which in total is $0.6$~km$^2$ big. Similar to the abstract world, we densely populate random objects on sidewalks to increase the difficulty. 
In total, we use $7$ towns and $14$ weather conditions for training and validation, and test the best policy on the real robot.

\noindent \textbf{Observation Space.}
Our AlienGo robot is equipped with the following onboard sensors: an RGB camera, a depth camera, a 1-D LiDAR, a GPS (from a cell phone), and a T265 camera for visual odometry. Additionally, we can generate semantic segmentation of the scene by processing RGB images with pretrained networks. Designing an observation space that enables sim-to-real transfer is a key contribution of this work and will be detailed in Section~\ref{sec:sim_to_real}. Here we list the observation modalities we use:

\begin{enumerate}

\item \textbf{Real Feature} (RFEAT) is a 128x60x80 feature tensor extracted from RGB image using pre-trained semantic network.

\item \textbf{Real LiDAR} (RLID) is a $272$ LiDAR information reading, representing the distance to the nearest object captured by casting a laser in all directions. The maximum sensing distance is capped at 6 meters.

\item \textbf{Real Goal direction \& distance} (RGDD) is the same relative goal location in the local polar coordinate frame, which is computed based on GPS signals.

\end{enumerate}

\begin{figure}
    \centering
    \includegraphics[width=\linewidth]{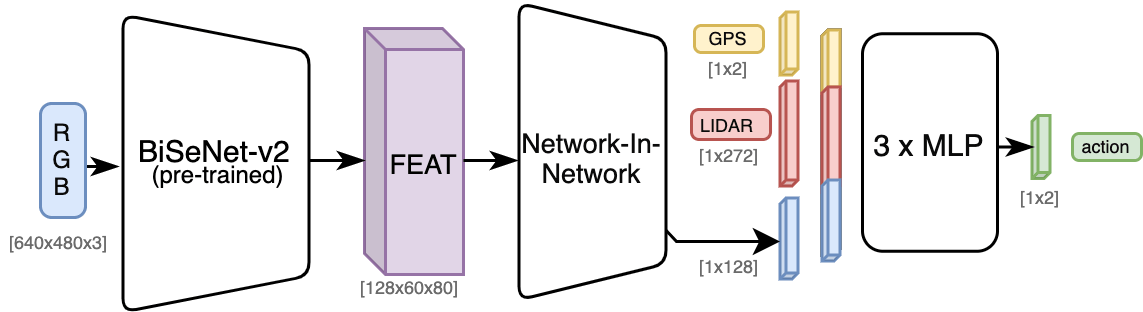}
    \caption{A network architecture of the student policy.}
    \label{fig:arch}
\end{figure}

\noindent \textbf{Behavior Cloning with DAGGER.}
We clone the teacher's behavior to the student policy using DAGGER.  DAGGER allows the student agent to learn proper actions in a supervised learning fashion in the presence of the expert. It generates a rollout using the current student policy, collects the corresponding \emph{abstract} and \emph{realistic} observations,
and updates the parameters by minimizing the L1 loss between the student and teacher actions. Note that in the high-fidelity simulator we can generate both \emph{realistic} and \emph{abstract} observations, where the latter is used by the teacher for generating labels (\figref{overview}). We bootstrap the learning by prefilling the replay buffer with $120k$ successful teacher experience pairs before the student training begins. The student architecture (\figref{arch}) consists of 3 components: pre-trained BiSeNet~\cite{yu2018bisenet} without the last layer for feature extraction, Network-In-Network~\cite{lin2013network}, and 3-layer MLP for feature processing. 

In our experience, our two-staged learning framework takes approximately $30$ hours to converge, while learning from scratch takes more than $300$ hours to converge.
\section{Sim-to-real transfer} \label{sec:sim_to_real}

\begin{table}[t]
\begin{center}
\begin{tabular}{lccc}

% \hline
\textbf{Agent}  &  \textbf{Train}  & \textbf{Valid}  & \textbf{Real World} \\
\hline

\textbf{TEACHER}  &  $79.60\%$  &  $83.18\%$  & $-$ \\
\hline

\textbf{ULTIMATE}  &  $67.68\%$  &  $76.86\%$  & $-$ \\
% \hline

% \textbf{LIDAR}  &  $38.62\%$  &  $56.26\%$  & $-$ \\
% \hline

\textbf{DEPTH}  &  $44.25\%$  &  $65.14\%$  & $-$ \\
% \hline

\textbf{GT-SEM}  &  $67.02\%$  &  $73.14\%$  & $-$ \\
\hline

\textbf{INF-SEM}  &  $55.01\%$  &  $63.26\%$  & $7.1\%$ \\
% \hline

\textbf{RGB}  &  \boldmath$77.92\%$  &  $73.16\%$  & $25.0\%$ \\
% \hline

\textbf{INF-SEM-FEAT (ours)}  &  $69.71\%$  &  \boldmath$77.90\%$  & \boldmath$83.3\%$ \\
\hline

\end{tabular}
\caption{
Success rate comparison of different agent configurations during training, validation, and real world testing. 
}

\label{table:all_agents_valid_set}
\end{center}
\end{table}

After learning in simulation, the next step is to deploy the policy to the real quadruped. While the \emph{learning by cheating} framework offers sample efficient learning, it can suffer from severe performance degradation during both behavior cloning and real-world transfer. To cross the \emph{sim-to-real} gap, we need to carefully design sensing modalities of the policy and data augmentation method to train the semantic model.

\subsection{Sensing Modalities}

In simulated environments, adding more sensing modalities often translates to simpler  learning problems~\cite{zhou2019does,kim2021observation}. However, in the real world, we need to cautiously select sensors because each sensor comes with its own \emph{sim-to-reap} gap. Therefore, we evaluate each sensor based on three criteria: (1) the usefulness of the information it encapsulates, (2) the additional difficulty in learning it adds, and (3) the sensitivity to the sim-to-real gap it induces.

To this end, we conduct an ablation study (Table~\ref{table:all_agents_valid_set}) of different agent configurations. Our simulation experiments help us to measure the usefulness (criteria 1) and learning difficulty (criteria 2), while real-world experiments evaluate sim-to-real transferability (criteria 3). 
For more details of real-world experiments, please refer to Figure~\ref{fig:real_avoidance_total} and Section~\ref{res:avoidance}.
Based on the experimental results, we select three sensor modalities: inferred semantic features (RFEAT), lidar (RLID), and goal direction and distance (RGDD). We will describe the selection process here.

\noindent \textbf{Sanity Check.} 
First, we conduct a sanity check to test whether the teacher's policy can be successfully transferred to the student agent with egocentric observations. For this purpose, we train an ideal \emph{ULTIMATE} agent that has access to all the sensors, raw RGB images, depth images, ground-truth semantic images, lidar, and localization. It achieves the success rate of $76.86\%$ that is only $6\%$ lower than the teacher (Table~\ref{table:all_agents_valid_set}). Therefore, we conclude it a successful sanity check.

\noindent \textbf{Visual Sensors.} Visual sensors, such as RGB, depth images, or semantics segmentations, are useful information to recognize surroundings. Table~\ref{table:all_agents_valid_set} indicates that all three agent with individual sensors, raw RGB images (\emph{RGB}), depth images (\emph{DEPTH}), ground-truth semantic images (\emph{GT-SEM}), show promising performance of $73.16$\%, $65.14$\% and $73.14$\%, respectively. However, ground-truth semantic is not available at deployment, and the performance drops to $63.26$\% when we infer it using a pre-trained semantic segmentation model (\emph{INF-SEM}). Then we investigate the third criteria, the sim-to-real transferability.
At this point, we stop investigating the \emph{DEPTH} agent because our infrared-based depth camera works poorly under direct sunlight in outdoor environments. Unfortunately, both \emph{INF-SEM} and \emph{RGB} agents show poor success rates of $7.1$\% and $25$\% in the real world. While the RGB agent seems a bit more promising, we are not able to improve its performance in the real world, after applying commonly-used sim-to-real techniques, such as domain randomization~\cite{laskin2020reinforcement}. 

The significant performance degradation of \emph{INF-SEM} in the real world is due to the poor generalization of the semantic segmentation model. Our insight is that the last layer of the model discards important features to classify each pixel into a small number of classes. Therefore, we suspect that the features before the output layer may contain more useful information than the final labels.

For this reason, we trained a new agent, \emph{INF-SEM-FEAT}, which takes the \emph{feature layer} (the last layer before prediction) of the semantic network as the input. This approach is common in transfer learning in computer vision~\cite{shah2021rrl}, but under-explored in visual navigation \cite{wijmans2019dd}. This agent achieves satisfactory success rates in both simulation and the real world, $77.90$\% and $83.3$\%, respectively. 
% \jie{It is unclear to me how these success rates are calculated, especially in the real world.} \sehoon{From Figure 3, right} \jie{I see. We need a forward pointer here then since Figure 3 and how it is generated are introduced in the next Section.}

\noindent \textbf{LiDAR.}
We select a LiDAR as an additional sensor because it contains useful distance information for collision avoidance and suffers less from noise in the real world. Therefore, we train all the agents with an additional LiDAR sensor.

\noindent \textbf{Localization.}
We test two common approaches for outdoor localization: visual odometry and GPS. In our experience, visual odometry, such as the Intel Realsense T265 sensor, is vulnerable to the accumulation of errors, and thus is not desirable for long-range navigation tasks. Therefore, we decide to use a GPS sensor on a Pixel5 phone, despite its high latency and noise levels.
We find that the phone's GPS works poorly when it is mounted on our legged robot. This is probably due to the high frequency vibrations of the robot's body, caused by periodic foot impacts during walking. 
Instead of using a more advanced GPS or developing a damping mechanism, which are orthogonal to the goal of this paper, we simply carry the phone with a human operator who walks closely with the robot.

\subsection{Data Augmentation}
The performance of semantic agents, \emph{INF-SEM} and \emph{INF-SEM-FEAT} highly depends on pre-trained semantic networks. However, when we tested publicly available segmentation networks, such as \cite{yu2018bisenet,mohan2020efficientps}, their performance is not satisfactory because they are mainly trained with autonomous-driving datasets and suffer from perspective shifts~\cite{shetty2019not} from road to sidewalk. Therefore, we take the BiSeNetV2 ~\cite{yu2018bisenet} implementation by CoinCheung ~\cite{BiSeNet_github} and slightly adapt it for our task.

To expand training set with additional sidewalk images, we create an augmented dataset that contains $17535$ sidewalk perspective and $9704$ road images synthesized using GTA ~\cite{richter2017playing} and $6000$ road and $2400$ sidewalk images synthesized using CARLA. We combine these newly generated synthesized images with the existing $2975$ real images of Cityscapes~\cite{cordts2016cityscapes} for training.

However, this data augmentation induces new sidewalk perspectives to the existing dataset, which previously had only road perspectives, and makes the learning of BiSeNet significantly more difficult. This is because the network tends to leverage the location of the surface as a strong prior. In other words, the original BiSeNet tends to predict the surface under the camera as a vehicle road. It is true for vehicle perspectives, but not true for sidewalk perspectives. 

We address this issue by solving a slightly different problem of segmenting the region a robot currently occupies. For example, if the robot is standing on the road, we segment the entire road in the field of view. Likewise, if the robot is currently walking on the sidewalk, the segmentation region will be the sidewalk.
Once trained, we evaluated the model on $35$ manually annotated images from the real robot's RGB camera and observed clear improvements.
\section{Experiments} \label{sec:experiments}

\begin{figure*}[ht]
    \centering
    \includegraphics[width=0.20\linewidth,height=5.2cm]{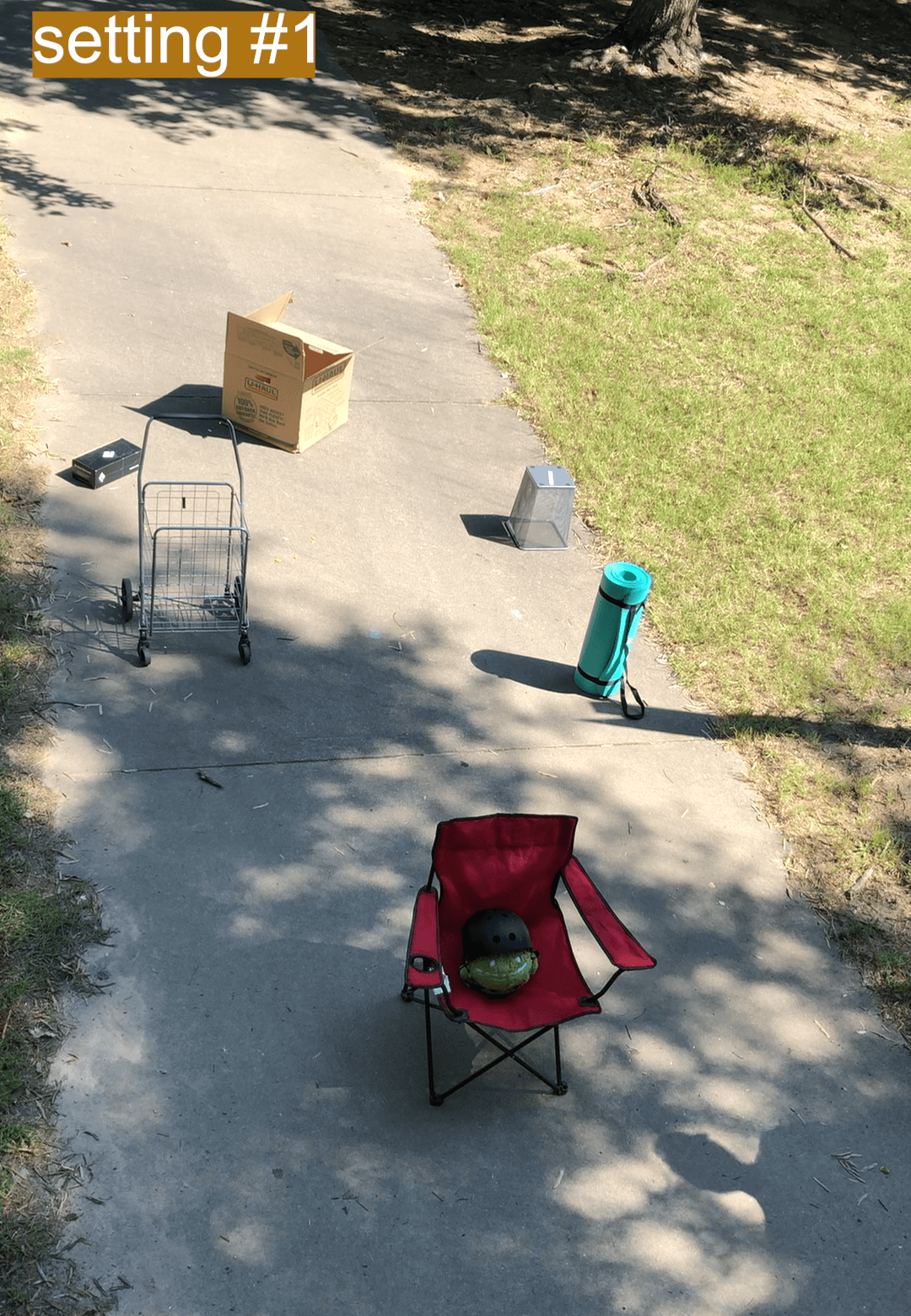} \hspace{-3mm}
    \includegraphics[width=0.20\linewidth,height=5.2cm]{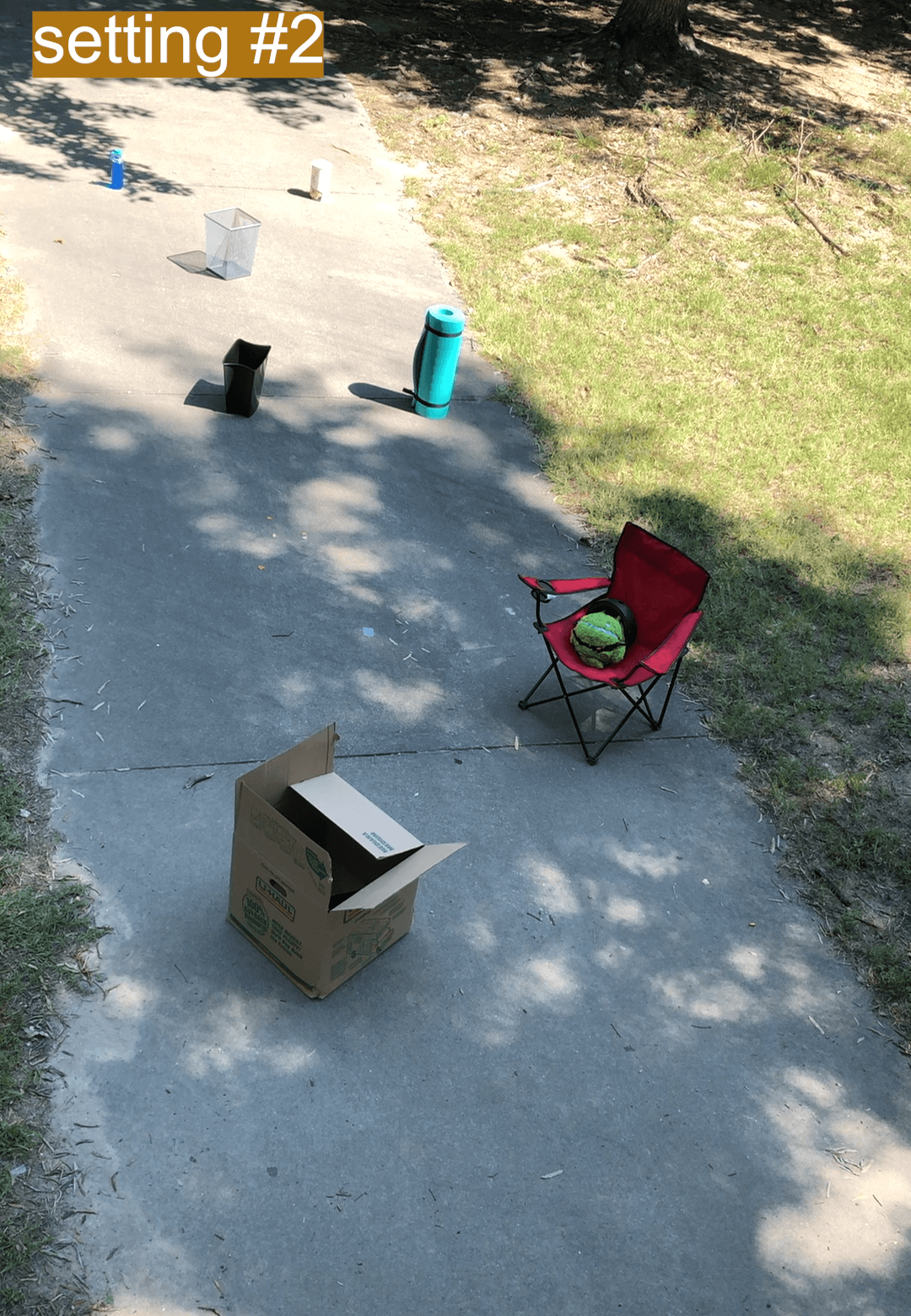} \hspace{-3mm}
    \includegraphics[width=0.20\linewidth,height=5.2cm]{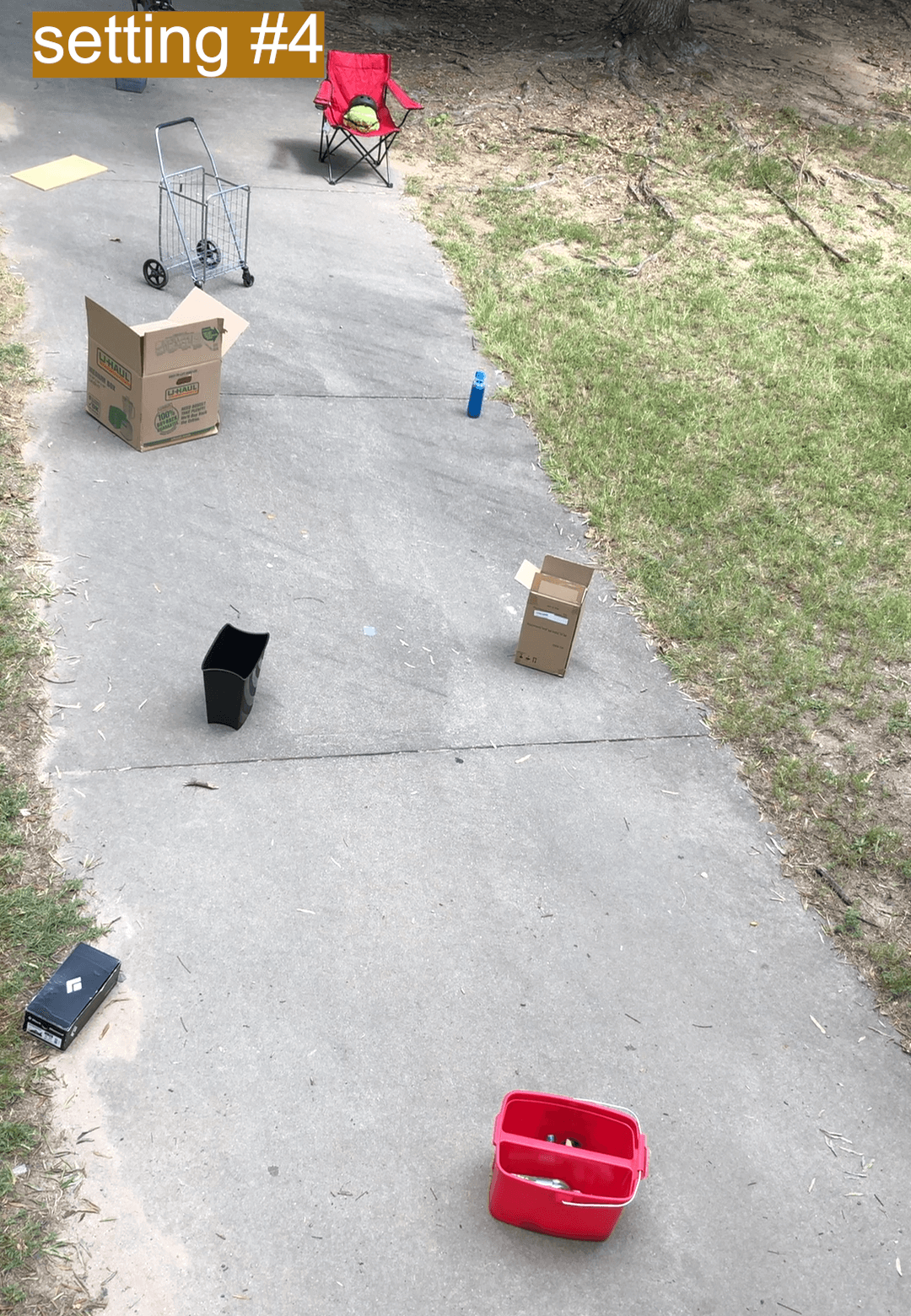} %\hspace{-3mm}
    \includegraphics[width=0.38\linewidth,height=5.22cm]{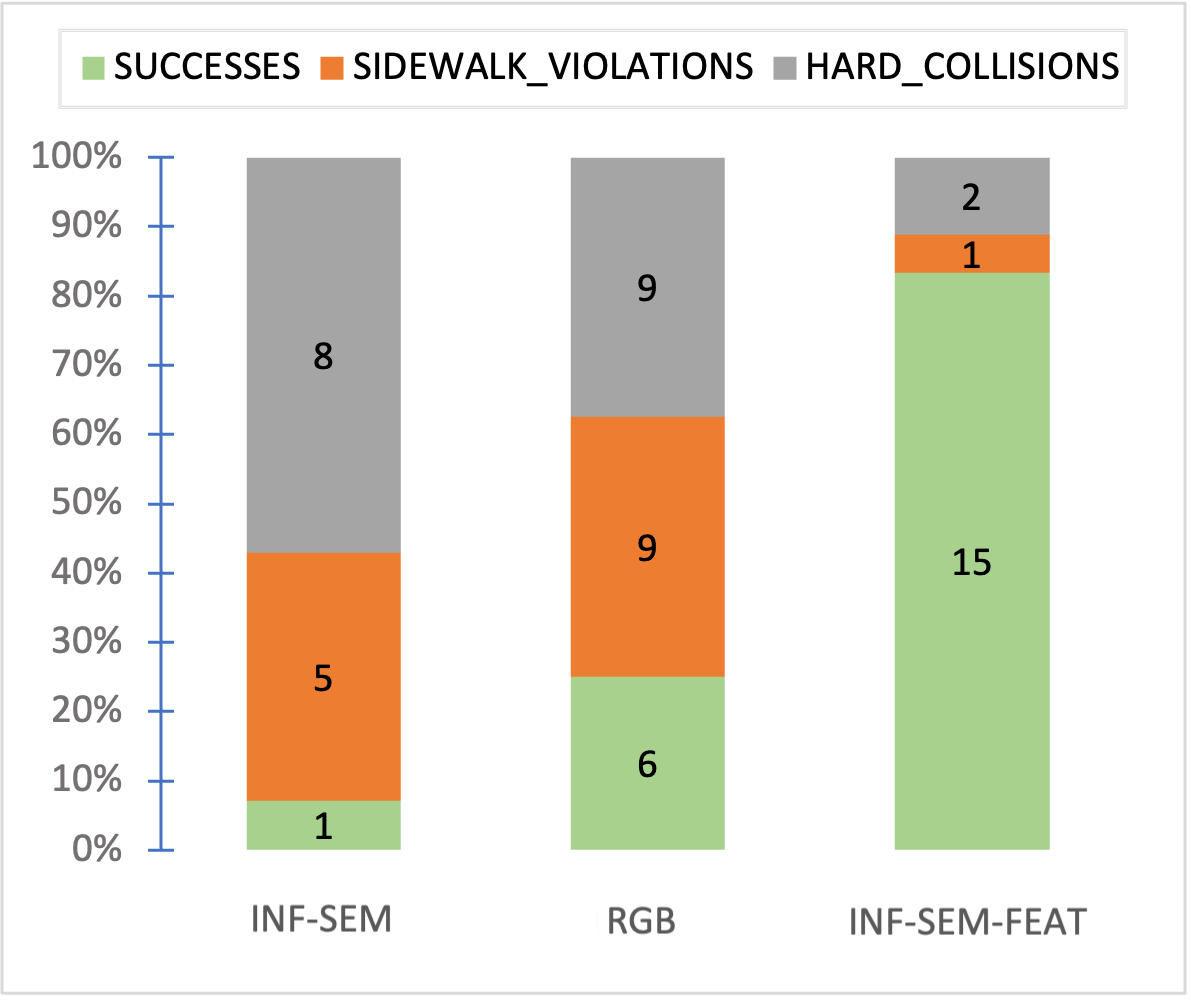} %\vspace{1mm}
    \caption{
    Comparison of the collision avoidance performance of three agents: \emph{RGB}, \emph{INF-SEM}, and \emph{INF-SEM-FEAT}.
    \textbf{Left:} Two out of four real world course configurations
    \textbf{Right:} Success rates with the breakdown of failure types: \emph{sidewalk violations} and \emph{hard collisions}. Our agent, \emph{INF-SEM-FEAT}, shows a much higher success rate compared to other two agents.
    }
    \label{fig:real_avoidance_total}

\end{figure*}

Three key requirements of navigating sidewalks are 1) staying on the sidewalk, 2) avoiding collisions with obstacles and pedestrians and 3) reaching the goal. To validate that our approach satisfies all three requirements, we perform two types of real-world experiments. In Section~\ref{res:avoidance}, we evaluate the ability of our method (\emph{INF-SEM-FEAT}) and two baselines (\emph{RGB} and \emph{INF-SEM}) on obstacle avoidance and sidewalk following. In Section~\ref{res:long_route}, we evaluate our best performing agent, \emph{INF-SEM-FEAT}, on long-range sidewalk navigation tasks, where the robot gets stress-tested in a wide variety of real-world scenarios. 

%%%%%%%%%%%%%%%%%%%%%%%%%%%%%%%%%%%%%%%%%%%%%%%%%%%%%%%%%%%%%%%%%%%%%%%%%%%%%%%%%%%%%%%%
\subsection{Obstacle Avoidance} \label{res:avoidance}

\noindent \textbf{Experiment Setup.}
We select a random sidewalk section of $15$ meters in length at the nearby park as a testing ground. We populate a variety of static obstacles represented that can be commonly found in a household (Figure~\ref{fig:real_avoidance_total} Left). Neither the sidewalk nor the objects were previously seen by the policy during training. 

The agent is instructed to walk along the sidewalk towards the other end. We disable GPS and the robot only relies on visual odometry for this experiment as the agent does not need to walk far and is only guided by directional information. The robot is placed at the starting location and runs until it triggers one of the termination conditions, judged by a human supervisor. Three possible outcomes are (1) \emph{success} when the agent successfully passes the obstacle course without any violation (2) \emph{collision} when the agent collides with obstacles and (3) \emph{sidewalk violation} when the agent steps out of the sidewalk. We ignore minor collisions that do not noticeably affect the robot's trajectory or obstacles' positions.

\noindent \textbf{Results.}
We test all three agents multiple times on four different sidewalk settings (Figure~\ref{fig:real_avoidance_total} Left). We report \emph{success rate} of each agent in Figure~\ref{fig:real_avoidance_total}. Right along with the breakdown of failure reasons: \emph{hard collisions} or \emph{sidewalk violations}.  In agreement with our finding in simulation, the \emph{INF-SEM-FEAT} achieves the highest success rate ($83.3\%$) when deployed in the real world. It only collides with obstacles only twice and violates the sidewalk once. the \emph{INF-SEM} and \emph{RGB} agents show much lower success rates, $7.1\%$ and $25.0\%$, respectively. We also observed that the \emph{RGB} agent shows a slightly lower \emph{hard collisions rate} than the \emph{INF-SEM} agent.

Both \emph{RGB} and \emph{INF-SEM} agents are more sensitive to variations in sunlight, as they are both attracted to well-lit regions resulting in sidewalk violation. In our experience, \emph{INF-SEM-FEAT} is more robust to variances in lighting conditions or textures. We conjecture that this is due to richer information of the feature layer compared to the thresholded semantic predictions or raw RGB pixels.

For minor collisions that did not trigger termination, we checked the recorded sensor data, and found that these collisions were often caused by sensor limitations, e.g. the object is behind the RGB camera or too small/thin for the LiDAR to detect. These collisions can be potentially mitigated by longer observation history or additional sensing.

%%%%%%%%%%%%%%%%%%%%%%%%%%%%%%%%%%%%%%%%%%%%%%%%%%%%%%%%%%%%%%%%%%%%%%%%%%%%%%%%%%%%%%%% 
\subsection{Long-distance Sidewalk Navigation} \label{res:long_route}

\begin{figure}
    \centering
    \vspace{-0.3cm}
    \includegraphics[width=0.99\linewidth]{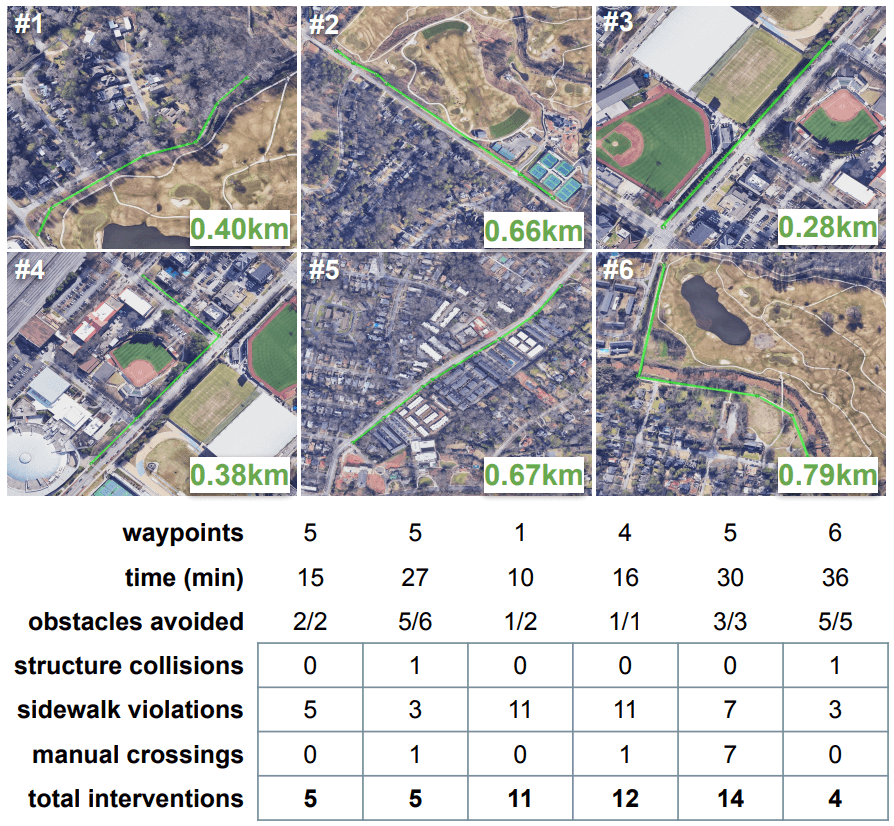}
    \vspace{-0.3cm}
    \caption{Our robot is able to autonomously navigate six routes using its onboard sensors, with limited human supervision.  We report a few statistics about the routes and the breakdown of interventions. 
    }
    \label{fig:long_track_stats}
\end{figure}

\noindent \textbf{Experiment Setup.}
To comprehensively evaluate our method, we deploy the learned navigation policy and the robot on a wide variety of real-world sidewalks and trails. We define six different routes in three locations around the city of Atlanta, USA, with total length of $3.2$~kilometers~(\figref{long_track_stats}). We select a set of GPS waypoints along each route. When the robot reaches within the $2$m radius of a given waypoint, it proceeds to the next one. 

\noindent \textbf{Interventions.} Human supervisors constantly monitor the robot and take over control in the following three cases:
\begin{enumerate}
    \item{Safety:} The robot walks too close to people around, or the robot gets into a dangerous situation for itself (\eg imminent collisions).
    \item{Sidewalk violation:} The robot walks out of the sidewalk boundary and does not automatically recover.
    \item{Road crossing:} The robot needs help crossing the road, as it is not yet trained to do so.
\end{enumerate}

\noindent \textbf{Results \& Discussion.} We summarize the numbers of how the policy performed on \figref{long_track_stats}. Over $3.2$~kilometer walk, the robot avoided 17 out of 19 obstacles. The longest non-interrupted section was $320$ meters, while the mean is from $31$m to $198$m, which varies to different courses.

\noindent \emph{Obstacle avoidance:}
The robot was exposed to naturally presented obstacles, such as poles, cones, and trash bins. Additionally, the human supervisors sometimes intentionally walked in front of the robot to pretend as pedestrians. The agent managed to avoid 17 out of 19 obstacles. The two collisions were against a trash bin and a narrow poll, which were minor and did not require human interventions. In both cases, the robot attempted to avoid the obstacle but collided with its side while passing around.

\noindent \emph{Sidewalk following:}
The agent generally possesses a good ability to stay on sidewalks. Although the robot sometimes failed at localization, it could still remain within the walkable regions. This robust behavior indicates that the agent is not only directed to the goal location but prioritizes staying in the sidewalk boundary. However, the agent sometimes still walked outside of sidewalk boundaries when it encountered visually challenging scenarios, such as bright regions or transitions to shadows.

Thanks to the rich features in the semantic model, the agent was able to navigate on different sidewalk surfaces, including pavement, tiled, and many others available in the simulation. However, we observed that certain types of terrains could cause confusions to the robot. For instance, we found that the agent was more prone to violate the sidewalk boundary on two particular routes. We analyzed these failure cases by reconstructing semantic images from the recorded features. It turned out that the agent misclassified grassy sidewalks as non-walkable areas. 

\noindent\emph{Localization Errors:}
In general, GPS performance was stable throughout the experiments. However, GPS occasionally experienced significant delays or noises, which caused errors in the localization, waypoint switching logic and ultimately led to sidewalk violations.

\noindent\emph{Driveways:}
Another main reason for manual interventions was the existence of driveways, a slope on the sidewalk to give car access to buildings. Note that driveways were not included in the training data. The agent often confused the semantics of the sidewalk and driveway due to the lack of data, and then proceeded in the direction of the driveway. Please refer to the supplemental video for examples. 

\begin{figure*}
    \centering
    \hspace{-1cm}
    \vspace{-0.5cm}
    \includegraphics[width=0.95\linewidth]{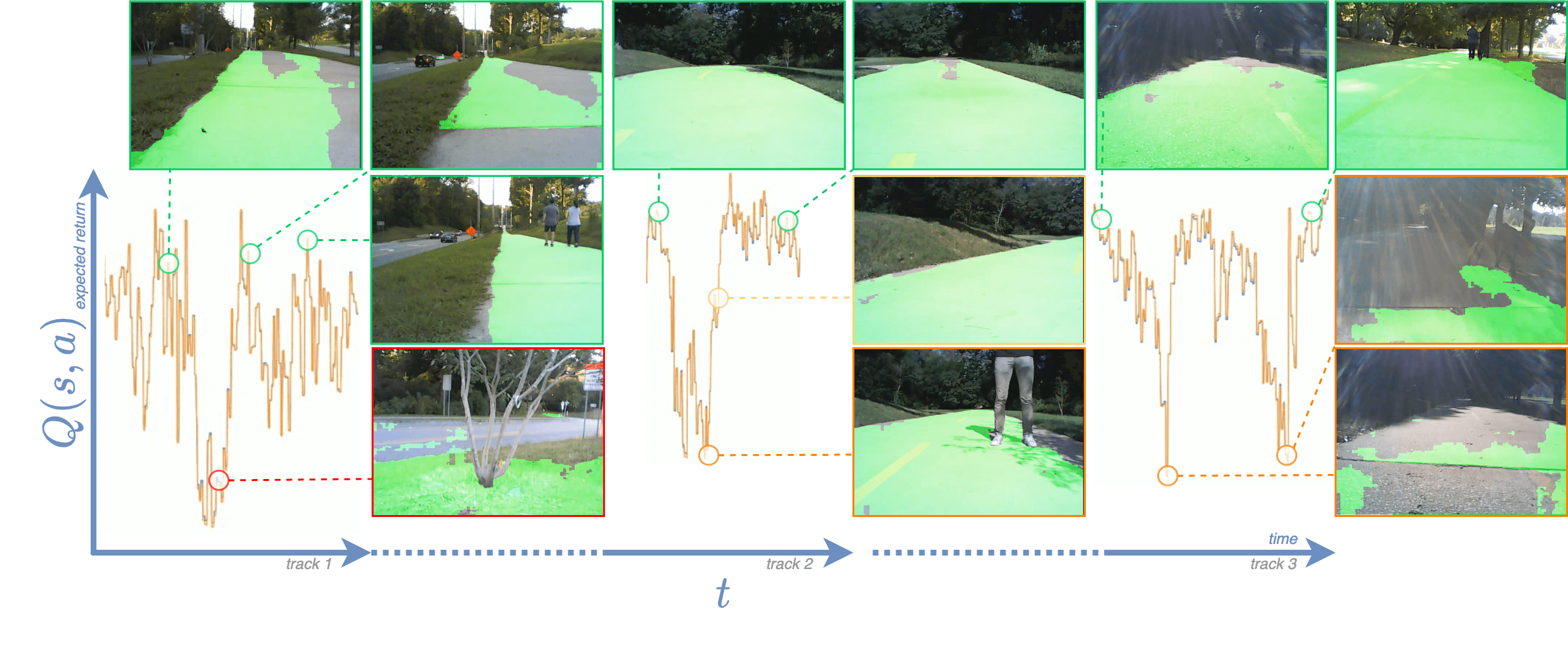}
    \caption{
    Q-function value predictions during real-world testing. Q-function values drop when an agent detects anomalies, such as sidewalk violation (\textbf{Left}), a human obstacle (\textbf{Middle}), and sun glare (\textbf{Right}), and recover back when the situation is resolved.
    }
    \label{fig:q_function}
\end{figure*}

\noindent \textbf{Q-function Analysis.}
Along with an actor, we also learn a Q function. When visualizing the Q function together with semantic maps, we find that the changes of Q-values are interpretable and intuitive. In \figref{q_function}, we illustrate a few interesting situations, such as sidewalk violation, encountering a pedestrian, and unclear vision due to sun glare. All these scenarios result in decreasing Q-values. Once the situations improve, the Q-values recover back to the normal range.
\section{Conclusion} \label{sec:conclusion}
We developed a quadrupedal robot that follows the route plan generated by the public map service, while remaining on sidewalks and avoiding collisions with obstacles and pedestrians using its onboard sensors. The main research effort focused on deploying a policy trained using a two-staged learning framework in simulation, to an unstructured and dynamic real outdoor environment. We discussed the important decisions on sensing modalities and presented new training procedures to overcome the sim-to-real gap. We evaluated our learning on a quadrupedal robot, which was able to walk 3.2 kilometers on sidewalks in the city of Atlanta, USA, with the learned navigation policy, onboard sensors, and a limited number of human interventions.

In the future, we plan to address a few limitations. First, the algorithm determines the waypoint transition moment based on an overly simplified rule-based logic that is vulnerable to localization errors. One potential approach is to leverage perceived semantics to determine transition moments, such as ``turning at the corner in front of the stop sign''. Second, the agent is subject to minor collisions when it passes the object due to the combination of limited sensing capability and lack of memory. In future work, we plan to address this issue by  investigating stateful models, such as recurrent neural networks. Finally, we want teach the robot dog to cross roads automatically, while it is 
now manually controlled by a human operator. This would require us to have traffic lights and crosswalks in simulation, as well as active control of vision~\cite{sorokin2021learning}.

\section*{Acknowledgements}
We would like to thank Visak Kumar, Nitish Sontakke, and Arnaud Klipfel for helpful discussions throughout the project, and their help during the real world deployment.

\bibliography{ref}
\bibliographystyle{IEEEtran}

\end{document}